\documentclass[preprint,11pt,nonatbib]{elsarticle}
\RequirePackage{fix-cm}
\RequirePackage[vmarginratio=1:1, outer=35mm, inner=25mm, lines=44]{geometry}
\usepackage{graphicx}
\usepackage{latexsym}
\usepackage[fleqn]{amsmath}
\usepackage{graphics}
\usepackage{algorithm}
\usepackage{algpseudocode}
\usepackage{graphicx}
\usepackage{multirow}
\usepackage{array}
\usepackage{booktabs}
\usepackage{amssymb}
\usepackage[dvipsnames]{xcolor}
\makeatletter
\@mathmargin\z@ 
\makeatother

\usepackage{csquotes}
\usepackage[%
colorlinks=true,
pdfborder={0 0 0},
linkcolor=green
]{hyperref}
\usepackage{multirow}
\usepackage{array}
\usepackage{booktabs}
\usepackage{caption}
\usepackage{subcaption}
\usepackage{setspace}
\makeatletter
\let\c@author\relax
\makeatother
\usepackage[
backend=biber,
style=numeric,
sorting=ynt
]{biblatex}
\usepackage{array}
\usepackage{makecell}

\begin{document}

\def\bsub#1{\def\theequation{#1\alph{equation}}\setcounter{equation}{0}}
\def\esub#1{\def\theequation{\arabic{equation}}\setcounter{equation}{#1}}

\def\subsubsection#1{\paragraph{\it #1}}







\begin{frontmatter}



\title{DenseBAM-GI: Attention Augmented DeneseNet with momentum aided GRU for HMER}


\author[inst1]{Aniket Pal\corref{mycorrespondingauthor}}
\cortext[mycorrespondingauthor]{First \& Corresponding author}
\ead{phc2017002@iiita.ac.in}

\author[inst1]{Krishna Pratap Singh}
\ead{kpsingh@iiita.ac.in}

\affiliation[inst1]{organization={Machine Learning \& optimization lab,Department of Information technology,
Indian Institute of Information Technology, Allahabad, India}}

\begin{abstract}

The task of recognising Handwritten Mathematical Expressions (HMER) is crucial in the fields of digital education and scholarly research. However, it is difficult to accurately determine the length and complex spatial relationships among symbols in handwritten mathematical expressions. In this study, we present a novel encoder-decoder architecture (DenseBAM-GI) for HMER, where the encoder has a Bottleneck Attention Module (BAM) to improve feature representation and the decoder has a Gated Input-GRU (GI-GRU) unit with an extra gate to make decoding long and complex expressions easier. The proposed model is an efficient and lightweight architecture with performance equivalent to state-of-the-art models in terms of Expression Recognition Rate (exprate). It also performs better in terms of top 1, 2, and 3 error accuracy ($ \leq 1(\%)$, $ \leq 2(\%) $ and $ \leq 3(\%)$) across the CROHME 2014, 2016, and 2019 datasets. DenseBAM-GI achieves the best exprate among all models on the CROHME 2019 dataset. Importantly, these successes are accomplished with a drop in the complexity of the calculation and a reduction in the need for GPU memory.
\end{abstract}


\end{frontmatter}


\section{Introduction}

HMER has recently attracted a lot of attention since it has potential uses in a number of industries, including conferencing systems, office automation, and education. Although machine learning and deep learning technologies have advanced quickly, the intricate spatial linkages and two-dimensional arrangements present in input pictures continue to provide a major challenge to developing efficient HMER solutions.

The process of HMER fundamentally involves symbol segmentation, symbol identification, and structural analysis, as discussed in various studies such as those by Zanibbi, Garain, and Alvaro et al (\cite{Zanibbi}, \cite{Garain}, \cite{Alvaro1}). Previously, strategies centred around pre-defined regulations and parsing techniques, as examined in the study by Anderson et al (\cite{10.1145/2402536.2402585_anderson}). Nonetheless, the advancements in HMER have been primarily driven by Deep Learning models, as evidenced in the works of Transformer and DenseWAP-TD ((\cite{DBLP:journals/corr/abs-2105-02412-transformer}, \cite{DenseWAP-TD})). Viewing HMER as an image-to-text problem within the mark-up context has proved successful, implementing deep learning's encoder-decoder structure by Deng et al ((\cite{DBLP:journals/corr/DengKR16})).
Following this, a number of encoder-decoder model modifications, such as  coverage-based encoder-decoder attention models, multiscale attention, and multi-modal  Subsequently, various adaptations of encoder-decoder models (\cite{zhang2017watch}, \cite{zhang2018multi}, \cite{multimodal}) have been developed. Furthermore, state-of-the-art performance has been achieved by Transformer-based and dual loss-based encoder-decoder models (\cite{dual-loss}, \cite{DBLP:journals/corr/abs-2105-02412-transformer}).

Despite achieving state-of-art (exprate upto 52\%), these methods exhibit limitations, including over-translation and under-translation issues (\cite{zhang2017watch}), an inability to capture intricate spatial relationships (\cite{PAL-V2}), substantial GPU memory requirements (\cite{PAL-V2}, \cite{DBLP:journals/corr/abs-2105-02412-transformer}), a lack of proficiency in representing length expressions (\cite{DenseWAP-TD}), and a necessity for improved generalization capabilities (\cite{DenseWAP-TD}).

Most of the deep learning research for HMER, is predominantly focused on developing the individual components of the encoder-decoder architecture, i.e., encoder or decoder. Architectures such as Fully Convolutional Networks (FCN) (\cite{zhang2017watch}) and DenseNet (\cite{zhang2018multi}, \cite{DenseWAP-TD}) are adopted in the encoder segment. However, these large-scale CNN models often encounter challenges related to gradient dynamics, such as exploding and vanishing gradients (\cite{resnet}, \cite{wideResNet}, \cite{shattered_gradients}). In order to mitigate these issues, attention mechanisms are integrated, originally proving their efficacy in areas like machine translation (\cite{DBLP:journals/corr/BahdanauCB14}, \cite{luong-etal-2015-effective}) and image classification (\cite{Residual_Attention}). In the mathematical images, attention is crucial to capture its 2D structure and complex spatial relations (\cite{zhang2017watch}). However, there is a noticeable deficiency in the academic discourse regarding integrating attention mechanisms within the encoder segment of encoder-decoder architectures applied to HMER models. In order to address this research gap, our present study introduces DenseBAM, an innovative model intended to amplify the representational capacity of the encoder. This proposed architecture melds the initial three blocks of DenseNet with the Bottleneck Attention Module (BAM), a fusion we designate as DenseBAM, to facilitate a more effective encoding process. BAM is a small integrable lightweight attention module proposed by Park et al. (\cite{DBLP:journals/corr/BAM}) and focuses on enhancing the network's representation power efficiently by introducing two attention mechanisms called channel \& spatial attention, which guides the network on ``what'' and ``where'' features should be emphasized. The channel attention mechanism captures the interdependencies in the channels of the feature map, and spatial attention captures the same in spatial features. Results show that it works best when applied in the high-level layers in the DenseNet.

The evolution of decoders in HMER models are marked by notable advancements with an initial application of Long Short Term Memory (LSTM) in combination with a CNN-based encoder proposed by Deng et al. (\cite{DBLP:journals/corr/DengKR16}). LSTM decoder later substituted with a more streamlined, two-layered stacked Gated Recurrent Unit (GRU), which provided similar performance but with fewer parameters. Subsequent innovation led to the development of a multimodal GRU-based decoder pioneered by Zhang et al. (\cite{multimodal}). Most recently, R-GRU (\cite{R-GRU}), a refined version of GRU,  emonstrated superior performance compared to the conventional GRU. Despite this progress, these models still grapple with challenges such as over-under parsing and less than satisfactory performance on long, complex mathematical expressions. In response to this necessity, we propose the GI-GRU, a novel model inspired by the momentum RNN (\cite{DBLP:journals/corr/abs-2006-06919_momentum_rnn}). This design integrates an auxiliary input into the naive GRU architecture, providing additional information that enhances the model's learning capacity. The integration of auxiliary input plays a crucial role in modulating the information update in the present hidden state derived from the current input. This integration not only fortifies the long-term memory retention capability of the GRU but also facilitates expedited convergence, thereby enhancing the overall efficiency of the model. Combination of proposed encoder and decoder, named as DenseBAM-GI, achieves state-of-the-art performance for top 1, 2, and 3 error accuracy for all HMER models on the CROHME datasets from 2014, 2016, and 2019. Similarly, it establishes a new benchmark for exprate on the CROHME 2019 dataset. Our novel contribution in this works are:  
\begin{enumerate}
    \item We propose a novel Encoder called DenseBAM which is equipped with channel and spatial attention mechanism.
    \item We develop a novel decoder named GI-GRU inspired from momentum based optimization techniques.
    \item The integrated encoder-decoder framework, named DenseBAM-GI, surpasses numerous leading models while reducing memory utilization and training duration.  
\end{enumerate}

\section{Literature review}
\label{SECRELATED}

Research in HMER primarily bifurcates into two methodologies: those that leverage rule-based approaches and those that employ encoder-decoder-based models.

Rule-based approaches were prevalent in deciphering mathematical expressions' two-dimensional structures, including the first research (\cite{10.1145/2402536.2402585_anderson}) in HMER. Zanibbi et al. (\cite{Zanibbi}) introduced the concept of expression grammar, an advanced form of Context-Free Grammar (CFG), to develop expression trees. It utilizes three stages: Layout, Lexical, and Expression Analysis. These stages methodically convert input images into LaTeX strings, optimizing this conversion process. Further research has focused on enriching this conversion by integrating classifiers and combining online and offline features, improving accuracy and overall performance. Furthermore, Yamamoto et al. (\cite{yamamoto:inria-00104743}) developed a Probabilistic CFG (PCFG) and it was extended as a stochastic CFG combining with HMM (\cite{Alvaro1}). Furthermore, Alvaro et al. (\cite{Alvaro2}) put forth a consolidated grammar-based approach to tackle the HMER challenge, and this method took first place in the CROHME 2014 contest.

Progress in deep learning-based models has presented a resort to grammar-based methods, enabling enhanced performance and the integration of automated segmentation and parsing. Deng et al. (\cite{DBLP:journals/corr/DengKR16}) proposed an encoder-decoder model incorporating a coarse attention mechanism consisting of a multi-layer CNN and LSTM decoder. Zhang et al. (\cite{zhang2017watch}, \cite{zhang2018multi}) further refined the model by integrating coverage-based attention, multi-scale attention, and stacked GRU layers. Wang et al. (\cite{multimodal}) unveiled a multimodal approach that strategically utilizes a multimodal encoder-decoder model. This innovative combination of online and offline modalities culminated in a substantial improvement in benchmark performance. Wu et al. (\cite{inbook}) further enhanced this strategy by formulating the Paired Adversarial Learning-v2 (PAL-v2) model. This system incorporated a Dense Convolutional Recurrent Neural Network (Conv-RNN) block, functioning as an encoder and replaced the traditional RNN decoder with an attention-based convolutional decoder (\cite{PAL-V2}). In a recent study, Zhao et al. (\cite{DBLP:journals/corr/abs-2105-02412-transformer}) unveiled an innovative bidirectionally trained transformer tailored for this particular domain. The uniqueness of this methodology is built on self-attention principles coupled with positional encodings.

Previous research has demonstrated the efficacy of Fully Convolutional Networks (FCNs) in extracting features from mathematical images (\cite{zhang2017watch}), with DenseNet later introduced as the encoder for efficient feature propagation (\cite{DenseWAP-TD}). Increasing the number of layers, as seen in VGGNet (\cite{vggnet}) and ResNet (\cite{resnet}), can enhance performance. However, deeper CNNs face challenges like exploding and vanishing gradients and vast parameter spaces. In response, the attention mechanism was introduced and gained prominence in various domains, offering minimal computational burden and significant performance improvement. Initially applied in Neural Machine Translation, attention was later integrated into CNNs for tasks like image classification. Channel-wise attention and adaptive strategies, such as the Bottleneck Attention Module, have also been implemented to modulate features dynamically. In this study, we propose DenseBAM encoder, which incorporates the BAM in the first three blocks of The DenseNet and enhance the baseline model performance with very little computational overhead.

Recurrent Neural Networks (RNNs) have long been a cornerstone for sequence modelling tasks, with their initial application in encoder-decoder architectures attributed to Cho et al. (\cite{cho}) in 2014. Even with their widespread use, RNNs intrinsically suffer from the vanishing and exploding gradient phenomena (\cite{RNN_gradient}, \cite{DBLP:journals/corr/RNN_vanishing_gradient}), which impede their capacity to learn long-range dependencies within sequences. To mitigate these challenges, gated RNN variants, such as LSTM (\cite{LSTM}) and GRU (\cite{GRU}), were introduced, providing enhanced capabilities in preserving information from elongated sequences. However, despite these advances, LSTMs and GRUs grapple with residual vanishing gradient issues, underscoring the necessity for ongoing research and refinement of RNN-based models for sequence learning endeavours. In this study, we propose the GI-GRU method to address the vanishing gradient issue and effectively capture long and complex spatial relationships in handwritten mathematical images. This technique outperforms the naive GRU while converging more rapidly.

We amalgamate the two novel proposed encoder-decoder components and evaluated the model against the CROHME benchmark datasets, wherein it exhibited performance commensurate with state-of-the-art models.

\section{Architecture of the model}
\label{sec:1}

\begin{figure*}
\hspace*{-1.6cm}
\includegraphics[width=20cm,height=10.5cm]{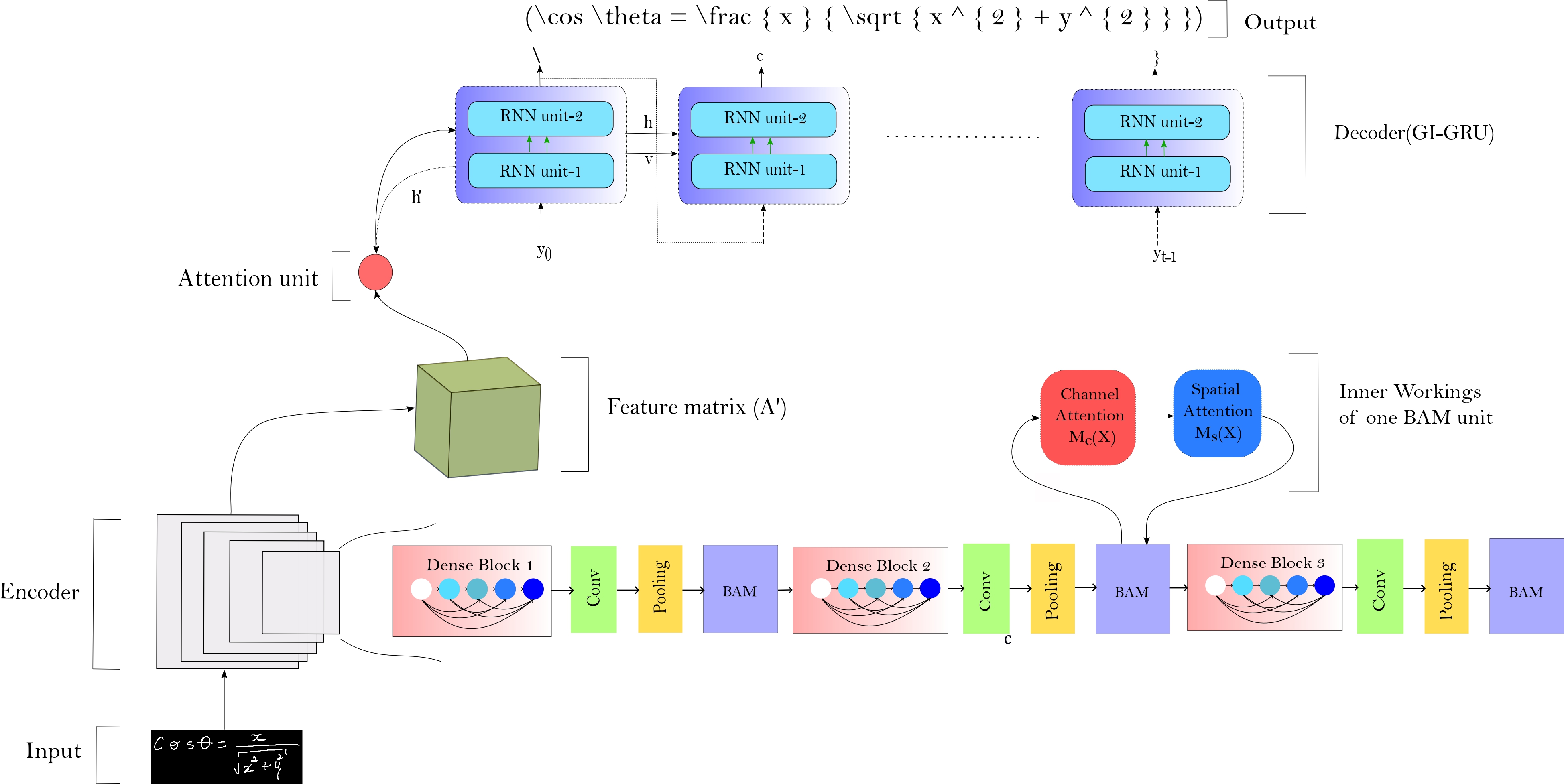}

\caption{In the architecture of our proposed model, the initial input constitutes an image of a handwritten mathematical equation. The embedded feature matrix, $\textbf{A}'$, is subsequently produced by the DenseBAM encoder we propose. Thereafter, the attention mechanism within the model identifies and emphasizes significant portions of the data, which it then conveys to the decoder unit. Within the decoder, the final output is generated by our newly proposed GI-GRU.
}

\label{Fig_Whole_architecture}
\end{figure*}

In our research, we employ an base encoder-decoder structure, enriched with attention mechanisms, as comprehensively expounded by Zhang et al. (\cite{zhang2017watch}). We refer to this particular architectural design as the base-model throughout this study. In our study, we introduce a novel encoder called DenseBAM, and the decoder module features a new variant of GRU, namely, GI-GRU. The proposed architecture aims to ensure quick convergence while capturing long-term dependencies. As depicted in Figure \ref{Fig_Whole_architecture}, the model's architecture encompasses an encoder, a decoder, and an attention module. It's important to note that this model undergoes end-to-end training in unison with all other components.

\subsection{Encoder}

The architecture we utilize as encoder incorporates the initial three Dense blocks from DenseNet-121. With its Fully Convolutional Network (FCN) type, it can adapt to various sizes of input images. This adaptability is paramount in HMER due to the variable length of equations. 
The encoder's role is to transform the provided input image into a form of feature representation, which is then subjected to further processing by an attention model-driven decoder. A robust feature representation is essential in HMER tasks, given the intricate relationships among mathematical symbols in the equation. Inadequate feature representation could lead to the subsequent module failing to generate an accurate Latex sequence corresponding to the image. We propose the DenseBAM encoder to bolster the power of feature representation by integrating the first three dense blocks with the BAM, generating a robust feature representation and substantially enhancing the model's performance.


{The architecture of the DenseBAM encoder is shown in Figure \ref{Fig_Whole_architecture}. It comprises the initial three dense blocks, each exhibiting a unique characteristic where the feature maps produced by a particular layer are concatenated with the features derived from all preceding layers.. This dense connectivity ensures that the information flows more efficiently and allows for better feature reuse. Consider an image passed through the convolution network, which comprises $L^{1}$ layers. A designated non-linear operation, termed $CF$, is executed within each layer, amalgamating Batch normalisation, ReLU transformation, pooling, and a convolution operation of $3 \times 3$. Thus, the output generated by the $l^{th}$ layer, symbolized as $q_{l}$, can be defined in the following manner: 
\begin{equation}
\begin{aligned}
q_l = CF_l([q_0, q_1, . . . , q_{l-1}])
\end{aligned}
\end{equation}

Here, the entity $[q_0, q_1, . . . , q_{l-1}]$ represents the concatenation of features emanating from a series of layers, specifically layers 0, 1, through to $l-1$. The number of layers and the dimension rise as the feature maps get concatenated. After applying $L^{1}$ layers in Dense block further a Bottleneck Layer which consist of $1\times1$ Convolutional Layer and Transition Layer consist of $1\times1$ Convolutional Layer  and $2\times2$ Average Pooling Layer further gets applied. Let the output from the Transition Layer can be represented as the following equation:

\begin{equation}
    Q^1 = {\text{Denseblock}_1} (q_l)
\end{equation}

Then this output is passed through Bottleneck Attention Module which has 2 Attention Layer called:

(i) Channel Attention Layer
(ii) Spatial Attention Layer

The channel and Spatial Attention Layer are represented by $M_{ch}()$ and $M_{sp}()$ respectively.

The $M_{ch}()$ comprised of a batch normalization Multi Layer Perceptron (MLP) and a Average Pooling Layer it can be represented as

\begin{equation}
    M_{ch}(I) = BN(MLP(Avg Pooling(I)))
\end{equation}

here I is the input to Channel Attention Layer.

Similarly we represent Spatial Attention Layer as $M_{sp}()$ and it consist of Convolutional Layer and Batch Normalization and can be represented as 

\begin{equation}
    M_{sp}(I) = BN(f_{1\times1}^3(f_{3\times3}^2(f_{3\times3}^1(f_{1\times1}^0(I)))))
\end{equation}

$f_{1\times1}^m$ and $f_{3\times3}^m$ are $1\times1$ and $3\times3$ Convolution and m is the index of convolution.

The $Q^1$ is passed $M_{ch}()$ and $M_{sp}()$ and the entire operation can be described as

\begin{equation}
    M(Q^1) = \sigma(M_{ch}(Q^1) + M_{sp}(Q^1))
\end{equation}

This $M(Q^1)$ is further added to $Q^1$ by elementwise multiplication and described as

\begin{equation}
    (Q^{\small{1}})^{'} = Q^{\small{1}} + Q^{\small{1}} \otimes M(Q^{\small{1}})
\end{equation}

This is the output of the $1^{st}$ dense block with BAM and input of $2^{nd}$ dense block. The final output of the DenseBAM Network is $\textbf{A}'$, has $L^1(H \times W)$ grids of features and in our study the $L^1$ is 1024. Each component in $l$ is characterized by an $N'$-dimensional vector, providing a detailed representation of a localized region within the image.

\begin{equation}
    a = \{a_1, ..., a_{L^{1}}\}, a_i \in \mathbb{R}^{N^{'}}
\end{equation}

\subsection{Attention Mechanism}
The use of attention enables the encoder-decoder structure precisely align the input and output sequences. Specifically, additive attention combined with convolution, as described by \cite{DBLP:journals/corr/cov-additive}, is employed. The approach has proven its efficiency across diverse applications. It is frequently used in work requiring complex sequence transformations, such as language processing, machine translation using neural networks, and other applications. This attention weight prioritizes important regions in the input for generating the next token in the output sequence. The attention component generates the context vector $c_t$ from the inputs $\textbf{A}'$ and $h^{'}_t$. The formula used to calculate it is provided by:
\begin{equation}
\begin{aligned}
\beta_t &= \sum_{l}^{t-1}\alpha_l;F = Q * \beta_t \\
e_{ti} &= \mathbf{\nu}^T_{a}\mathbf{tanh}(W_{h'}h'_t + W_aa_i + W_ff_i + b_i); \\
\alpha_{ti} &= \frac{exp(e_{ti})}
{\sum_{k=1}^{L}exp(e_{tk})};
c_t = \sum\alpha_{ti}*a_{i}
\end{aligned}
\end{equation}

where $i$ represents the position in the matrix $\mathbf{A{'}}$ and $t$ stands for the current timestamp. $\beta_t$ is a feature of the attention aggregation. The $i^{th}$ component of the coverage vector $F$ is symbolised by $f_i$, and the $Q$ denotes the convolution layer equipped with $q$ output channels. The attention sum vector is introduced into the convolution layer $Q$ to produce the coverage vector $F$. Additionally, $e_{ti}$ measures the energy of $a_i$ at the particular timestamp $t$. The first RNN cell's and encoder's outputs are $h^{'}_t$ and $a_i$, respectively. The attention weight of the feature map $a_i$ at $t$ is $\alpha_{ti}$. The second RNN cell then receives the context vector $c_t$ to compute $h_t$.

Additionally, we have $\nu \in \mathbb{R}^{d'}$, $W_{h'} \in \mathbb{R}^{d' \times n}$, $W_a \in \mathbb{R}^{d' \times N}$, and $W_f \in \mathbb{R}^{d'\times q}$, where $d'$ denotes attention dimension.

\begin{figure*}
\includegraphics[width=17cm,height=5.5cm]{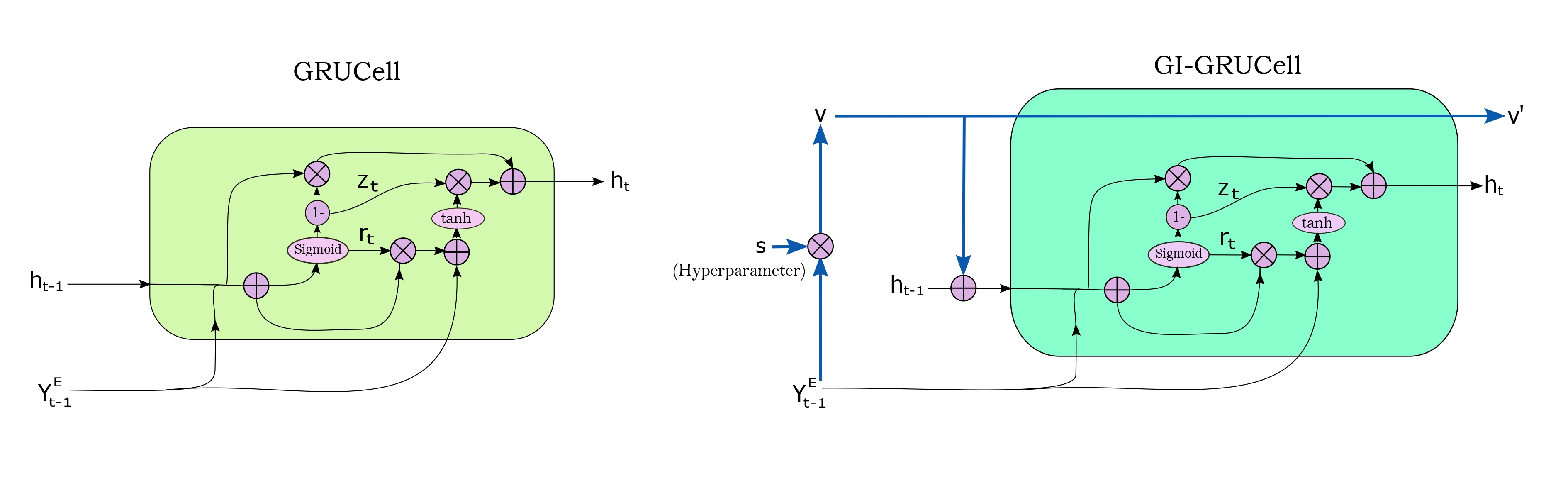}

\caption{Architecture of proposed GI-GRU Cell.
}

\label{Fig_GRU}
\end{figure*}

\subsection{Decoder}
Two layers of stacked RNN architecture made up the decoder unit. An intermediate hidden state is produced by the first layer, which is then used as the input for the subsequent recurrent neural network (RNN) and attention unit. The final hidden state, the context vector, and the current output word are all produced by the second layer. We employ the GRU as the RNN unit for decoding. It has fewer parameters and is effective for addressing the vanishing gradient problem. In this paper, a novel decoding unit called GI-GRU is developed to produce a sequence of Latex by taking use of the interaction between the context vector and the feature representation matrix.

Figure \ref{Fig_GRU}(a) illustrates the structure of a GRUCell. The mathematical representation of $h_t$ in GRU is:

\begin{equation}
\label{GRU}
\begin{aligned}
z_t &= \sigma (W_{yz}y^E_{t-1} + U_{hz}h_{t-1} + C_{cz}c_t + b_z) \\
r_t &= \sigma (W_{yr}y^E_{t-1} + U_{hr}h_{t-1} + C_{cr}c_t + b_r) \\
\tilde{h}_t &= tanh(W_{yh}y^E_{t-1} \\
&+ r_t \otimes (U_{rh}h_{t-1}) + C_{ch}c_t + b_h) \\
h_t &= (1 - z_t) \otimes h_{t-1} + z_t \otimes \tilde{h}_t
\end{aligned}
\end{equation}

where the symbol $\sigma$ symbolizes the sigmoid function, while $\otimes$ denotes element-wise multiplication. $y^E_{t-1}$ refers to the previously timestamped predicted word or label incorporated with its embedding. The variables $z_t$ and $r_t$ represent the update and reset gates, respectively. Moreover, $\tilde{h}_t$ indicates the candidate activation, while $h_t$ designates the hidden state at the $t^{th}$ instance.

\par

\subsubsection{GI-GRU}
We introduce an auxiliary state $v_t = s\otimes W_{yv}y^E_{t-1}$, which is added to three elements, namely, $z_t$ (update gate), $r_t$ (reset gate), and $h_t$ (current hidden state), by drawing on the principles of the classical momentum technique, which is renowned for its capacity to accelerate model convergence and mitigate the vanishing gradient issue.
The model we propose is christened as GI-GRU, with its definition as follows:

\begin{equation}
\label{GI-GRU}
\begin{aligned}
v_t &= s\otimes W_{yv}y^E_{t-1} \\
z_t &= \sigma (W_{yz} (y^E_{t-1}) + U_{hz}h_{t-1} + v_t + C_{cz}c_t + b_z) \\
r_t &= \sigma (W_{yr} (y^E_{t-1}) + U_{hr}h_{t-1} + v_t + C_{cr}c_t + b_r) \\
\hspace*{-1cm}\tilde{h}_t &= tanh(W_{yh} (y^E_{t-1}) \\
& + r_t \otimes(U_{hh}h_{t-1} + v_t) + C_{ch}c_t + b_h)\\
h_t &= (1 - z_t) \otimes h_{t-1} + z_t \otimes \tilde{h}_t
\end{aligned}
\end{equation}

where $s$ is the Step size parameter. The variable $v_t$ adeptly modulates the information updated to the gates, thus expediting the convergence of the Gated Recurrent Unit (GRU) and enhancing its capability to retain longer sequences. It is achieved by fostering a more stable dynamical system and alleviating vanishing gradient. The Gated Input GRU (GI-GRU) is correctly titled since the gate only introduces the input to the GRU. The Algorithm 1 describes computing processes.

\subsection{Loss Function of DenseBAM-GI}

The calculation of the target LaTeX sequence for a given timestep, denoted as $t$, can be expressed as follows:

\begin{equation}
\label{softmax}
\begin{aligned}
\mathrm{P}(y_t|y_1,..,y_{t-1},x)  &=  g(W_{o}o_t), \\
&= g\left(W_o\left(\mathrm{V_{h}h_{t} + W_{c}c_{t} +  W_{y}y^{E}_{t-1}}\right)\right),
\end{aligned}
\end{equation}
	
where $\mathrm{W_o \in \mathbb{R}^{K \times w'}, V_h \in \mathbb{R}^{w'\times n}, W_c \in \mathbb{R}^{w'\times N}}$ and
$W_{y} \in {\mathbb{R}^{w'\times E}}$, $g$ signifies the softmax activation function, $K$ is indicative of the aggregate count of words present within the vocabulary, while $n$ embodies the dimensionality of the hidden state. Additionally, $\mathrm{w'}$ signifies the dimension of the model's linear layer. To minimize overfitting and ensure generalizability, the model employs a loss function based on cross entropy and imparts L2 regularization to the model's weights. Hence, the optimization goal can be described by the following objective function:

\begin{equation}
\begin{aligned}
\mathrm{F = - \sum_{t=1}^{S}\log{P}(y_t|y_{t-1},x) + \lambda||\textbf{W}||^2}
\end{aligned}
\end{equation}

where $S$ denotes the length of the output sequence, $x$ refers to the training data comprised of images of handwritten mathematical expressions, $\textbf{W}$ represents the model's parameters, and $\lambda$ is indicative of the hyperparameter. 

\begin{algorithm}
\caption{Computational steps of DenseBAM-GI }\label{alg:cap}
\begin{algorithmic}[1]

\State \textcolor{red}{Input}: HMER Image
\vspace{0.5cm}
\State \textcolor{red}{Encoder}: The encoder processes a batch of images representing handwritten mathematical expressions and subsequently generates a corresponding matrix of features, denoted as $\mathbf{A^{'}}$.
\vspace{0.5cm}
\State \textcolor{red}{Attention unit}: It creates the context vector $c_t$ using the inputs $\mathbf{A^{'}}$, the current attention sum $\beta_t$, and the hidden state of the previous timestamp. 
\vspace{0.5cm}
\State \textcolor{red}{Decoding}: Two level stacked GI-GRUCells are used for decoding purpose.   

\Statex \textbullet~{First level GRU is}:

\vspace{0.25cm}
 
 $v^{'}_t, h^{'}_t =\textbf{GI-GRU}(y^E_{t-1},h_{t-1},v_{t})$ 
 \vspace{0.25cm}
 
  \hspace{-0.8cm} where the inputs to first are true label or predicted word ($y^E_{t-1}$), hidden state ($h_{t-1}$), auxiliary state ($v_{t}$) created at the timestamp $t$ by this expression $s\otimes W_{yv}y^E_{t-1}$.
\vspace{0.25cm}

\Statex \textbullet~Functional form of Second level GRU can be written as

\vspace{0.25cm}
 
  $v_t, h_t =\textbf{GI-GRU}(c_{t},h^{'}_{t},v^{'}_{t})$

\vspace{0.25cm}

\hspace{-0.680cm} In this level GRU, inputs are context vector($c_{t}$), hidden and momentum auxiliary states ($h^{'}_{t}$,$v^{'}_{t}$) generated by first level GRU.
\vspace{0.5cm}

\State  \textcolor{red}{Output}: The output at a given timestamp, denoted as $\mathrm{y_t}$, is derived from the parameters $h_t$, $y^{E}{t-1}$, and $c_t$ in
$\mathrm{P(y_t|y_1,..,y_{t-1},x)  =  g(W_{o}o_t)}$
(eq-12).
\vspace{0.5cm}

\end{algorithmic}
\end{algorithm}

\section{Experimental Design}
\label{Exp}

The proposed DenseBAM-GI model's performance is assessed in this study using the CROHME 2014, 2016, and 2019 datasets. Each inkml file in these datasets contains x and y coordinates for online data traces. Each inkml file is converted into a binary image as part of the data preparation for the model. The whole model training and testing dataset is created by pairing these images with the corresponding labels.There are 101 classes in the training data, which includes 8,836 expressions from the CROHME 2014 and 2016 datasets as well as 1,157 expressions from the CROHME 2019 dataset. In comparison to the CROHME 2014, 2016, and 2019, test datasets, which each had 986, 1,136, and 1,199 novel expressions, respectively. The official CROHME 2014 dataset only serves as the training set for the suggested model. Performance evaluations were carried out utilising an 11 GB Nvidia RTX 2080 Ti graphics card.

The Stochastic Gradient Descent (SGD) approach with a momentum coefficient of 0.9 was used to train the models, and the training procedure was continued for 300 iterations. The learning rate was set to 0.0001 at the beginning and used an update approach that reduced it to the current rate/10 if performance did not increase for ten consecutive epochs. For L2 regularisation, the hyper-parameter lambda ($\lambda$) was fixed at 0.01. We used pre-trained weights from naïve GRU to initialise the GI-GRU model in order to promote quick convergence. We used the exprate and the Word Error Rate (WER) metrics to evaluate the performance of the suggested models.

Overall, this study provides a rigorous evaluation of the proposed models using standardized datasets, providing valuable insights into their effectiveness in recognizing handwritten mathematical expressions.

\begin{table*}[hbt!]
		\centering
		\caption{Comparison of the proposed model with the cutting-edge models on the CROHME 2014 test sets.}
		\resizebox{0.8\textwidth}{!}{%
		\begin{tabular}[t]{l | c | c | c | c }
			\toprule
			\thead{Models}& Exprate (\%) 
 & $\leq 1(\%)$ & $\leq 2(\%)$ & $\leq 3(\%)$\\
			\midrule
			I \cite{crohme2014} & 37.22 & 44.22 & 47.26 & 50.20  \\
			II \cite{crohme2014} & 15.01 & 22.31 & 26.57 & 27.69 \\
			IV \cite{crohme2014} & 18.97 & 28.19 & 32.35 & 33.37 \\
			V \cite{crohme2014} & 18.97 & 26.37 & 30.83 & 32.96 \\
			VI \cite{crohme2014} & 25.66 & 33.16 & 35.90 & 37.22  \\
			VII \cite{crohme2014} & 26.06 & 33.87 & 38.54 & 39.96  \\
	    
	   WYGIWYS \cite{DBLP:journals/corr/DengKR16} & 28.70 & - & - & - \\
	        WAP \cite{zhang2017watch}& 40.04 & 56.1 & 59.9 & -  \\
	        WAP(with ensemble) \cite{zhang2017watch}& 44.40 & 58.40 & 62.20 & 63.10  \\
	    	End-to-end \cite{End-to-End-2017}& 25.09 & - & - & -  \\
			PAL \cite{zhang2018multi} & 39.66 & - & - & -  \\
			PAL-v2 \cite{PAL-V2} & 43.81 & - & - & - \\
			PAL-v2(with printed data) \cite{PAL-V2} & 48.88 & 64.50 & 69.78 & 73.83 \\
			DenseWAP-TD \cite{DenseWAP-TD} & 49.1 & 64.2 & 67.8 & - \\
			Transformer(uni) \cite{DBLP:journals/corr/abs-2105-02412-transformer} & 48.17 & 59.63 & 63.29 & - \\
			Transformer(bi) \cite{DBLP:journals/corr/abs-2105-02412-transformer} & \textbf{\fbox{53.96}} & 66.02 & 70.28 & - \\
            DenseWAP-CTC \cite{DenseWAP_CTC}  & 50.96 & - & - & - \\
            {R-GRU}\cite{R-GRU} & {43.72} & - & - & - \\
            AdamR-GRU(\cite{AdamR-GRU}) & 50.32 & 68.39 & 75.97 & 82.25 \\
			\textbf{base-model} & 43.11 & 62.31 & 70.77 & 77.03 \\
            \textbf{with DenseBAM} &53.50  & 70.17 & 78.23 & 83.54 \\
            \textbf{DenseBAM-GI} &51.69  & \textbf{\fbox{70.27}} & \textbf{\fbox{78.87}} & \textbf{\fbox{83.75}} \\
			
			\bottomrule
		\end{tabular}%
}
\end{table*}

\begin{table*}[hbt!]
		\centering
		\caption{Comparison of proposed model with the state-of-the art models on CROHME 2016 test sets.}
		\resizebox{0.8\textwidth}{!}{%
		\begin{tabular}[t]{l | c | c | c | c }
			\toprule
			\thead{Models}& Exprate (\%) 
 & $\leq 1(\%)$ & $\leq 2(\%)$ & $\leq 3(\%)$ \\
			\midrule

	   Tokyo \cite{crohme2016} & 43.90 & 50.91 & 53.70 & - \\
	   sao paolo \cite{crohme2016} & 33.39 & 43.50 & 49.17 & - \\
	   Nantes \cite{crohme2016} & 13.34 & 21.02 & 28.33 & - \\
	        WAP \cite{zhang2017watch}& 37.1 & - & - & -  \\
	        WAP(with ensemble) \cite{zhang2017watch}& 44.55 & 57.10 & 61.55 & 62.34  \\
			PAL-v2 \cite{PAL-V2} & 43.77 & - & - & - \\
			PAL-v2(with printed data) \cite{PAL-V2} & 49.61 & 64.08 & 70.27 & 73.50 \\
			DenseWAP-TD \cite{DenseWAP-TD} & 48.5 & 62.3 & 65.3 & - \\
                DenseWAP-CTC \cite{DenseWAP_CTC}  & 49.96 & - & - & - \\
			Transformer(uni) \cite{DBLP:journals/corr/abs-2105-02412-transformer}  & 44.95 & 56.13 & 60.47 & - \\
			Transformer(bi) \cite{DBLP:journals/corr/abs-2105-02412-transformer} & \textbf{\fbox{52.31}} & 63.90 & 68.61 & - \\
            R-GRU\cite{R-GRU} & {41.29} & - & - & - \\
            AdamR-GRU(\cite{AdamR-GRU}) & 47.68 & 64.48 & 75.46 & 81.36 \\
			\textbf{base-model} & 42.28 & 58.62 & 71.05 & 78.22 \\
    
			\textbf{with DenseBAM} &49.54  & 65.42 & 76.91 & 83.21 \\
            \textbf{DenseBAM-GI} &51.18  & \textbf{\fbox{67.33}} & \textbf{\fbox{77.28}} & \textbf{\fbox{83.66}}   \\
			\bottomrule
		\end{tabular}%
}
\end{table*}

\begin{table*}[hbt!]
		\centering
		\caption{Comparison of proposed model with the state-of-the art models on CROHME 2019 test sets.}
		\resizebox{0.7\textwidth}{!}{%
		\begin{tabular}[t]{l | c | c | c | c }
			\toprule
			\thead{Models}& Exprate (\%) 
 & $\leq 1(\%)$ & $\leq 2(\%)$ & $\leq 3(\%)$\\
			\midrule

	   Univ.Linz\cite{CROHME2019} & 41.49 & 54.13 & 58.88 & - \\
        TUAT\cite{CROHME2019} & 24.10 & 35.53 & 43.12 & -\\
	        WAP \cite{zhang2017watch}& 41.7 & 55.5 & 59.3 & -  \\
	        DenseWAP-TD \cite{DenseWAP-TD} & 51.4 & 66.1 & 69.1 & - \\
			Transformer(uni) \cite{DBLP:journals/corr/abs-2105-02412-transformer}  & 44.95 & 56.13 & 60.47 & - \\
			Transformer(bi) \cite{DBLP:journals/corr/abs-2105-02412-transformer} & 52.96 & 65.97 & 69.14 & - \\
            AdamR-GRU(\cite{AdamR-GRU}) & 50.00 & 65.82 & 73.27 & 80.35 \\
		\textbf{base-model} & 39.59 & 58.78 & 68.85 & 73.89  \\
            \textbf{with DenseBAM} &50.88  & 68.34 & 77.42 & 82.09 \\
            \textbf{DenseBAM-GI} &\textbf{\fbox{52.99}}  & \textbf{\fbox{71.34}} & \textbf{\fbox{79.36}} & \textbf{\fbox{84.21}}   \\
			\bottomrule
		\end{tabular}%
}
\end{table*}

\section{Results and Discussion}

This discourse is organized into three segments. First, the proposed model is juxtaposed with the state-of-art models using the benchmark datasets from CROHME 2014, 2016, and 2019 to evaluate comparative performance. Following this, within our proposed model, an ablation study is performed to scrutinize the individual contributions of the BAM and GI-GRU. The final segments of this section are dedicated to analyzing the exprate relative to different sequence length and comparing our proposed model's computational complexity with that of other prominent models in the field.

\begin{table*}[hbt!]
		\centering
		\caption{Results of applying BAM in DenseNet with 6,12,24-layers at each denseblock with naive-GRU.}\label{tab1}
            \vspace{0.2cm}
		\resizebox{0.9\textwidth}{!}{%
		\begin{tabular}[t]{l | c | c | c | c }
			\toprule
			\thead{BAM in different positions} & Exprate  & $\leq 1(\%)$ 
                     & $\leq 2(\%)$ & $\leq 3(\%)$  \\
			\midrule
			1. After denseblock1 &47.55  &64.65 & 74.09 &79.72 \\
                2. After denseblock2 &diverges  &--- &--- &--- \\
                3. After denseblock3 &51.06  &67.51 &76.85 &81.63 \\
                4. After denseblock1 and denseblock3 &47.22 &62.82 &72.65 &78.52 \\
                5. After denseblock2 and denseblock3 &53.50 &70.17 &78.23 &83.54 \\
                6. After denseblock1 and denseblock2 & 42.99 & 59.44 &70.17 &77.07 \\
                7. After each denseblock &46.39 &65.67 & 73.77 &79.51 \\

			\bottomrule
		\end{tabular}%
}
\end{table*}

\begin{table*}[hbt!]
		\centering
		\caption{Results of applying Constant and variable size convolution layers in the DenseBAM encoder (with naive GRU) on CROHME 2014 test dataset. }\label{tab1}
            \vspace{0.2cm}
		\resizebox{0.95\textwidth}{!}{%
		\begin{tabular}[t]{l | c | c | c | c }
			\toprule
			\thead{Encoder Architecture (Number of convolutional layer in each DenseBlock)} & Exprate  & $\leq 1(\%)$ 
                     & $\leq 2(\%)$ & $\leq 3(\%)$  \\
			\midrule
                1. DenseBAM (12,12,12) & Diverges  & - & - & -\\
                2. DenseBAM (16,16,16) & 47.32  & 66.56 & 75.10 & 81.30 \\
                3. DenseBAM (6,12,24) &53.50  & 70.17 & 78.23 & 83.54 \\

			\bottomrule
		\end{tabular}%
}
\end{table*}

\begin{table*}[hbt!]
		\centering
		\caption{Results of applying BAM without and with momentum in CROHME 2014 test dataset. }\label{tab1}
            \vspace{0.2cm}
		\resizebox{\textwidth}{!}{%
		\begin{tabular}[t]{l | c | c | c | c }
			\toprule
			\thead{Architecture} & Exprate  & $\leq 1(\%)$
                     & $\leq 2(\%)$ & $\leq 3(\%)$  \\
			\midrule
			1. DenseNet + Naive GRU (base-model) &43.11  & 62.31 & 70.77 & 77.03 \\
                2. DenseNet + BAM + Naive GRU (DenseBAM with naive GRU ) &53.50  & 70.17 & 78.23 & 83.54 \\
                3. DenseNet + BAM + GI-GRU (DenseBAM-GI) &51.69  & 70.27 & 78.87 & 83.75 \\

			\bottomrule
		\end{tabular}%
}
\end{table*}
\subsection{An assessment of the proposed model in comparison to the most recent state-of-the-art techniques.}

An extensive comparison analysis is done on the CROHME 2014, 2016, and 2019 datasets to validate the efficacy of the proposed models (DenseBAM-GI). Only those systems that used official CROHME training data and did not make use of ensemble techniques were taken into consideration for a fair comparison. The top 1, 2, and 3 error accuracies as well as the exprate are also documented. The most recent comparison of the CROHME 2014 dataset is shown in Table 1. The performance of the participating models, I through VII, was assessed based on the results of the CROHME 2014 competition. Our proposed model, DenseBAM-GI, outperforms all models except the transformer model, with recognition rates of 51.19\%. However with scores of 70.17\%, 78.23\%, and 83.54\% for the top 1, 2, and 3 error accuracy, the proposed models set a new benchmark.

On the CROHME 2016 dataset, DenseBAM-GI models attain expression rates of 51.18\% percent. The most recent comparison of the models on the CROHME 2016 dataset is shown in Table 2. Top 1, 2, and 3 error accuracy scores for the DenseBAM-GI model were 67.33\%, 77.28\%, and 83.66\%, respectively. This performance sets a new standard in the field by outperforming all previous models in terms of top 1, 2, and 3 error accuracies.

On the CROHME 2019 dataset, the DenseBAM-GI outperforms all competing systems with an expression rate of 52.99\%. Table 3 shows the results with current state-of-the-art models. DenseBAM-GI achieves top 1, 2, and 3 error accuracies of 71.34\%, 79.36\%, and 84.21\%, respectively. 

\subsection{Ablations studies}

In this section, we perform ablation studies to thoroughly examine the proposed model, focusing on the effects of applying the BAM and incorporating an additional auxiliary gate in the GRU. Moreover, we investigate the performance of BAM under varying configurations of convolutional layers within the DenseBlock, aiming to identify the optimal settings.
In addition to these studies, attention visualization is provided to facilitate a comprehensive analysis of BAM's influence on the attention-learning process, shedding light on its role in enhancing the model's performance.

\begin{figure*}
\hspace*{-1.4cm}
\includegraphics[width=20cm,height=19cm]{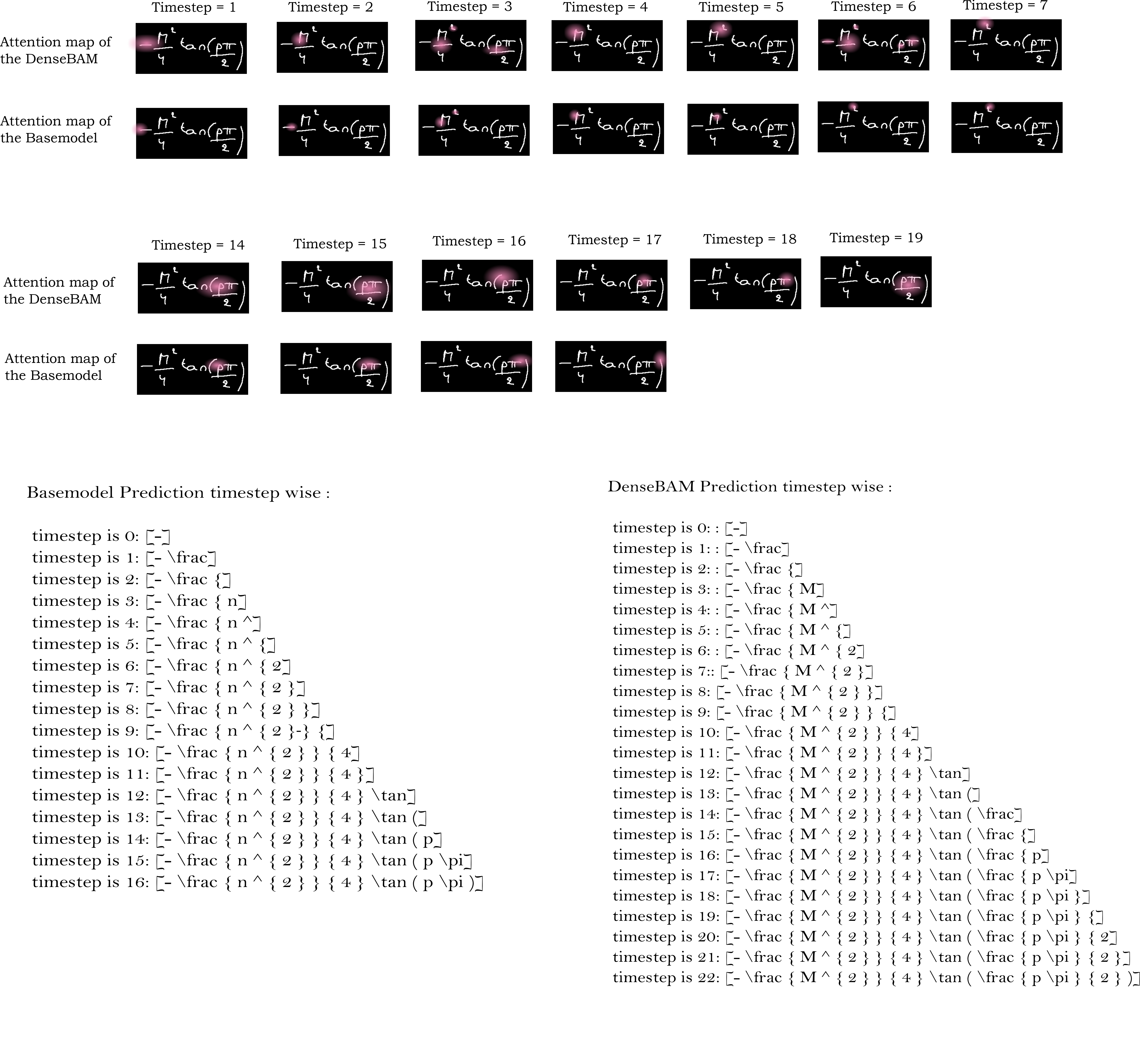} 

\caption{ Recognition result comparison between proposed DenseBAM encoder (with naive GRU Decoder) and the base-model on different image of expression.}

\label{attention_compare_expresssions}

\end{figure*}

\subsubsection{Effect of applying BAM in different bottleneck positions of DenseNet Encoder}

To attain optimal performance, it is crucial to appropriately position the Bottleneck Attention Module (BAM) within the dense blocks in DenseNet. In the study conducted by Park et al. (\cite{DBLP:journals/corr/BAM}), the BAM was positioned following each residual block. In our investigation, we examine the performance of BAM when placed after individual dense blocks and in conjunction with others. The most favourable results were obtained when BAM was placed after Denseblock 2 and 3 with naive GRU decoder, achieves exprate of 53.50\%. Table 4 demonstrates these findings. Additionally, when positioned after Denseblock 3 exclusively, the exprate reached 51.03\%. Based on these observations, it can be deduced that the optimal placement of BAM is within high-end layers responsible for capturing high-level features. These layers grasp the critical discriminative characteristics, necessitating attention to select the most pertinent information.

\subsubsection{Effect of constant and variable size conv layers in DenseNet with BAM}

The DenseNet architecture comprises a varying number of convolutional layers within each DenseBlock. In our model, we utilize the first three DenseBlocks from DenseNet-121. Some studies (\cite{DBLP:journals/corr/abs-2105-02412-transformer}) have applied a constant number of convolutional layers in each DenseBlock, prompting us to conduct experiments to ascertain the optimal settings for BAM. Table 5 shows the results, with the positions of BAM in DenseNet remaining consistent with the previous section.

Initially, we experiment by setting 12 convolutional layers (each layer contains $1 \times 1$ and $3 \times 3$ conv layers as in Table 1 of \cite{Densenet}) uniformly across each dense block, followed by increasing this number to 16. Finally, we reverted to the original structure of DenseNet. In the first scenario, the model diverged, while in the second case, the model achieved a 47.50\% exprate. However, enhanced results were obtained when we returned to the original configuration, applying 6, 12, and 24 convolutional layers in the DenseBlocks.

These observations suggest that, for optimal BAM performance within DenseNet, it is advantageous to use fewer convolutional layers at the beginning and more towards the end, as high-level features are predominantly located in the latter layers of the DenseNet architecture.

\subsubsection{Effect of Combining GI-GRU decoder with DenseBAM encoder}

In the conducted experiments, the DenseBAM encoder was paired with both a naive GRU decoder and a GI-GRU decoder. The baseline model only attained an exprate of 43\%. However, integrating the BAM boosted the exprate to 53.50\%. Table 6 depicts this results. Despite this improvement, the performance remained subpar on the CROHME 2016 and 2019 datasets (Table 2 \& 3), and the top 1, 2, and 3 error accuracies remained less than ideal. The incorporation of the GI-GRU decoder remedied these shortcomings, with the model achieving state-of-the-art results in top 1, 2, 3 accuracies, and also obtaining state-of-the-art exprate on the CROHME 2019 dataset. This improvement can be attributed to the auxiliary gate of the GI-GRU decoder, which assists in the retention of longer sequences, mitigates over-parsing and under-parsing issues, and fortifies the robustness of the language model.

\begin{figure*}
\hspace*{-1.8cm}
\includegraphics[width=21cm,height=6.5cm]{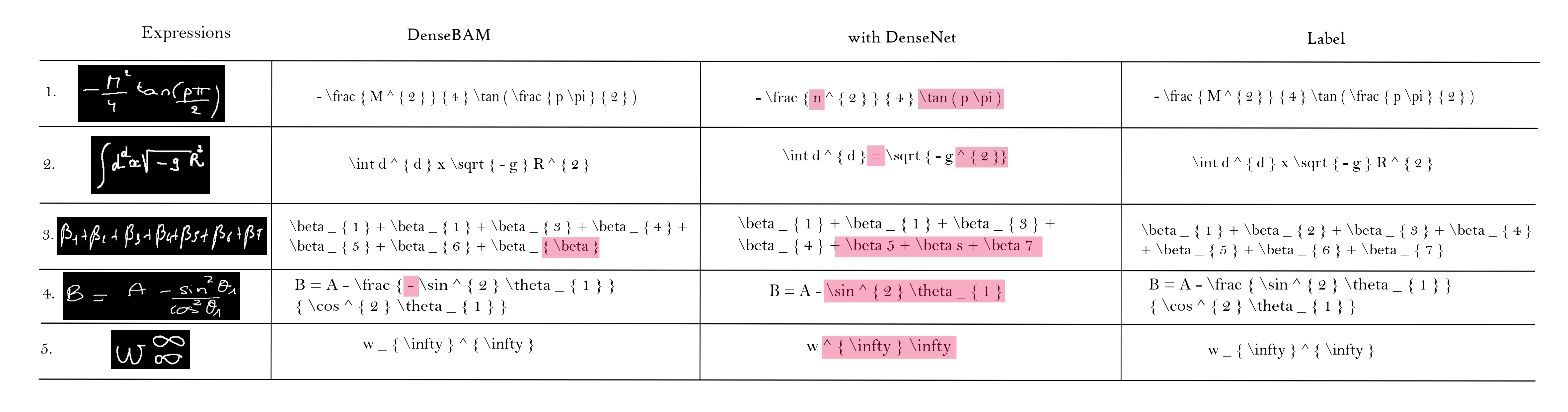} 

\caption{ Recognition result comparison between the proposed DenseBAM Encoder and Densenet Encoder (both with naive GRU decoder) and the base model on different image of expression.}

\label{Fig_compare_expresssions}

\hspace*{-2.0cm}
\includegraphics[width=21cm,height=6.5cm]{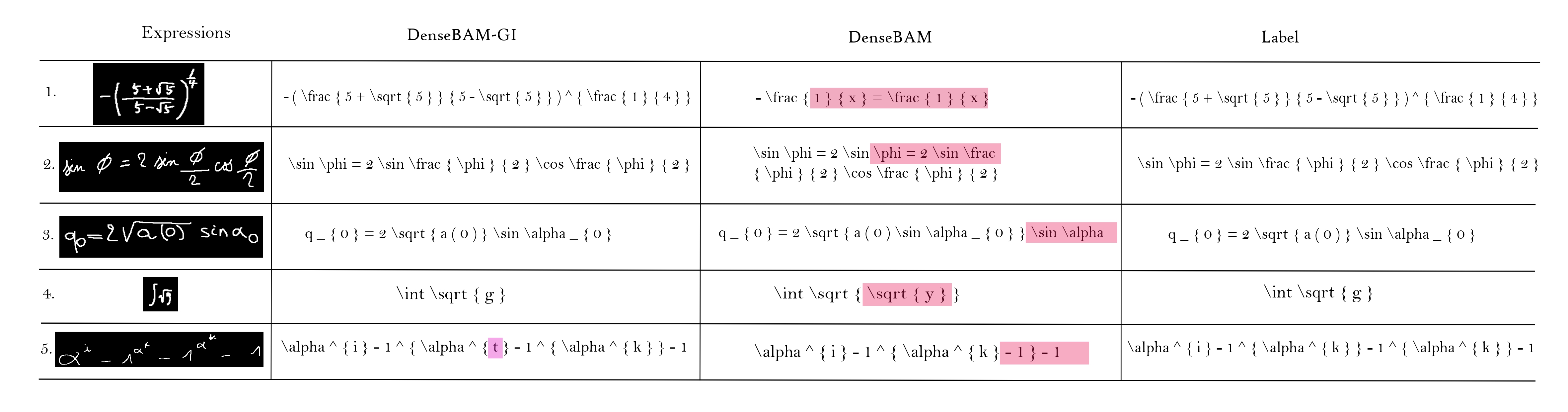} 

\caption{ Recognition result comparison between proposed DenseBAM Encoder (with naive GRU decoder) and proposed DenseBAM-GI model on different image of expression.}

\label{Fig_compare_expresssions_1}
\end{figure*}

\subsubsection{Attention visualization with BAM and without BAM}

To elucidate how the BAM improves the encoder's feature representation and facilitates recognition, we generate attention maps for both DenseBAM encoder (with naive GRU decoder) and the base-model, as illustrated in Figure 1. We examine attention maps at crucial timesteps, where BAM plays a vital role in predicting the LaTeX sequence.
In the first row, the first seven timesteps demonstrate that DenseBAM accurately recognizes the symbol `M', while the base model fails. Timesteps 2 and 4 are particularly critical for DenseBAM in identifying the symbol `M'. The attention areas for DenseBAM are notably larger and more distinct than the base model. Furthermore, DenseBAM focuses on both `frac' tokens in timestep three and again between timesteps 14 and 19. In contrast, the base model does not, resulting in its failure to recognize the second `frac' token.
These observations show that integrating the BAM attention technique in the encoder substantially alters the base model's attention mechanism, concentrating on the image's significant areas at the appropriate timesteps. Consequently, we deduce that BAM enhances the encoder's feature representation capabilities.

\subsubsection{Performance comparision at expression level between DenseNet and DenseBAM}

In a comprehensive analysis, we compare two encoders: DenseNet (the base encoder) and DenseBAM (the proposed encoder). The results are illustrated in Figure \ref{Fig_compare_expresssions}. The base model could not identify the relationship between the `frac' token and failed to recognize the `M' symbol. Moreover, the second expression did not capture the `X' symbol and its superscript relations, while DenseBAM successfully did so. In the subsequent two expressions, the base model failed to detect subscript and superscript relationships, leading to an incorrect LaTeX sequence. Although DenseBAM preserved all spatial relationships, it did not recognize the `7' symbol, and the `-' symbol was over-translated. The final mathematical expression is also misidentified due to the complex spatial relationships between superscripts and subscripts.

These findings suggest that the proposed DenseBAM model effectively captures most expressions with intricate spatial relationships among symbols, despite occasional over-translation. We attribute this performance to the additional attention mechanism within the encoder, which facilitates learning complex spatial relationships for the overall architecture. Furthermore, the results indicate an improvement in feature representation, as nearly all symbols are accurately identified.

\subsubsection{Performance comparision at expression level between DenseBAM and DenseBAM-GI}

In this section, a comparison is conducted between DenseBAM and DenseBAM-GI, focusing on the effect of integrating an additional gate into the GRU by appending BAM to the encoder at the expression level. Figure \ref{Fig_compare_expresssions_1} illustrates the results. In the case of the first expression, DenseBAM did not successfully capture the intricate spatial relationships between the `frac' and `sqrt' components, while DenseBAM-GI managed to preserve these relationships. DenseBAM over-translates the `sin' and `frac' tokens for the second expression, but DenseBAM-GI is corrected. In the third and fourth expressions, the `sin alpha' component was over-translated by DenseBAM, but DenseBAM-GI mitigated this issue. The last expression, which is lengthy and features a limited number of complex spatial relationships among superscripts, presented an under-translation challenge for DenseBAM, but DenseBAM-GI effectively addressed this problem.
The findings from this comparison indicate that most errors associated with DenseBAM stem from over-translation and under-translation. However, incorporating an additional gate in DenseBAM-GI alleviates these issues. It can be inferred that the extra gate in DenseBAM-GI strengthens the language model, facilitating the decoder in generating precise LaTeX representations for intricate and lengthy mathematical expressions.

\begin{figure*}
\hspace*{-2cm}
\includegraphics[width=20cm,height=7.5cm]{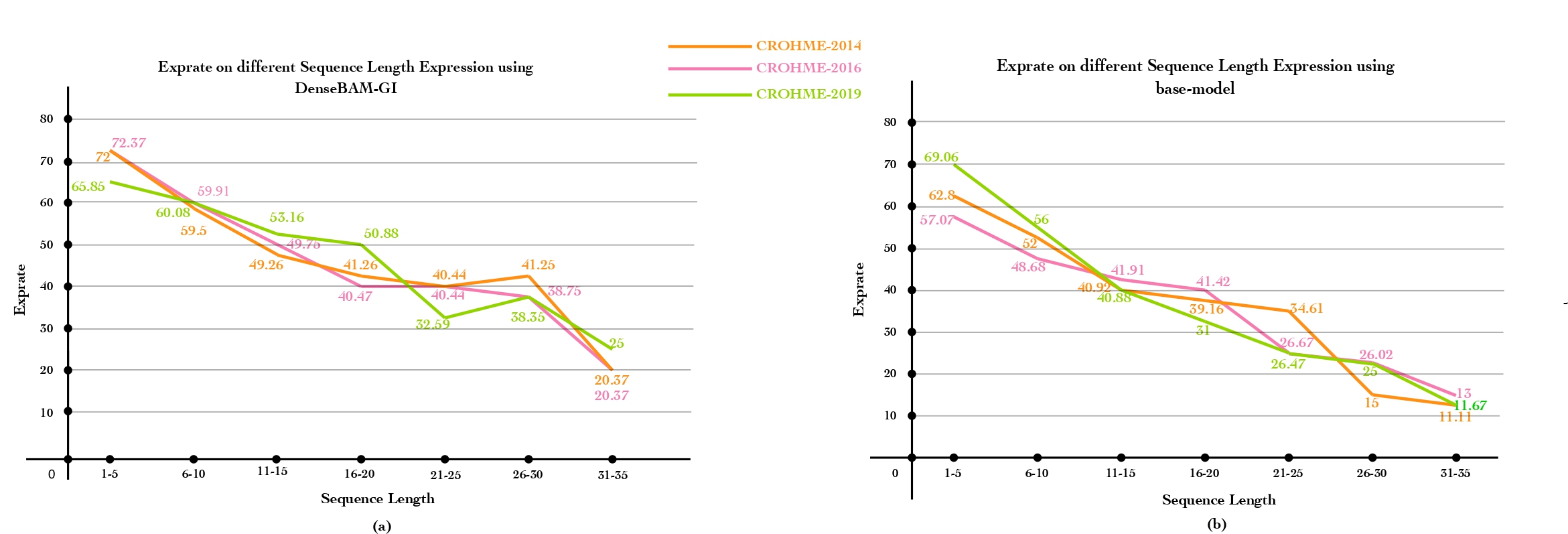} 
\caption{Comparative Analysis of DenseBAM-GI and Base-Model Performance Across Varied Expression Lengths Within CROHME 2014, 2016, and 2019 Datasets: (a) Assessment of DenseBAM-GI performance across disparate expression lengths, (b) Evaluation of base-model performance across varying expression lengths.}

\label{sequence_length}
\end{figure*}

\begin{table}[htb]
\centering
\caption{Performance of DenseBAM-GI \& base-model on different sequence length on CROHME 2014 test data}
\label{1234}
\resizebox{0.75\textwidth}{!}{%
\begin{tabular}{c*{6}{>{$}c<{$}}}\toprule
     \text{length} &
     \text{Exprate 
     with DenseBAM-GI} & \text{Exprate with base-model} & \text{Training data}            \\
    \midrule
1-5 \hspace{0.5cm}  & 72 & 62.8 & 985 \\
6-10 \hspace{0.5cm}  & 59.5 & 52 & 754 \\
11-15 \hspace{0.5cm}  & 49.26 & 41.91 & 939 \\
16-20\hspace{0.5cm}  & 41.26 & 39.16 & 675 \\
21-25\hspace{0.5cm}  & 40.44 & 34.61 & 761 \\
26-30\hspace{0.5cm}  & 21.67 & 15 & 465 \\
31-35\hspace{0.5cm}  & 16.21 & 13 & 534 \\
\bottomrule

\end{tabular}
}
\end{table}

\begin{table}[htb]
\caption{Comparison of the computational complexity of DenseBAM-GI.}
\label{1234}
\resizebox{0.5\textwidth}{!}{%
\begin{tabular}{c*{1}{>{$}c<{$}}}\toprule
     \text{System name} &
     \text{Computational Complexity}            \\
    \midrule
\text{PAL-V2 (\cite{PAL-V2})} \hspace{0.5cm}  & O(knd^2*nd^2 + knd^2) \\
\text{Transformer (\cite{DBLP:journals/corr/abs-2105-02412-transformer})} \hspace{0.5cm}  & O(knd^2) + O(L(2n^2d + 2ndd_{ff})) \\
\text{WAP (\cite{zhang2017watch})} \hspace{0.5cm}  & O(5(nd^2 + knd^2))\\
\text{DenseBAM-GI (Proposed model)} &
O(n(d^2 +c') + n(C^2 + (h*w)) + knd^2)\\
\bottomrule

\end{tabular}
}
\end{table}

\subsection{ The influence of input sequence length on the exprate of the proposed model 
}

For a more comprehensive understanding of the exprate, a detailed comparison between the DenseBAM-GI and the base model is conducted, focusing on the influence of sequence length variability on different mathematical expressions. As shown in Table 7 and Figure \ref{sequence_length}, which includes separate plots for the DenseBAM-GI and base models, the proposed DenseBAM-GI model demonstrated superior performance compared to the base model (naive GRU-based) across all sequence length categories. The figure also compares the exprate on CROHME 2016 and 2019 test datasets with CROHME 2014. Notably, the DenseBAM-GI model consistently registered an Exprate 3-10\% higher than the base model.
In the CROHME 2014 training dataset, consisting of 8836 expressions, 986 are short series (1-5). As the sequence length increased, the quantity of data decreased correspondingly, with long mathematical expressions (26-30) comprising roughly half the quantity of shorter sequences. Even though only 465 such lengthy expressions were available, the DenseBAM-GI model achieved an Exprate of 21\%, in contrast to the base model's 16\%. The complexity of symbol relationships in lengthy expressions poses significant recognition challenges for models, yet the proposed model significantly outperforms the base model (by more than 3\% in Exprate) in this area.
In addition, we observe that the proposed DenseBAM-GI maintains an accuracy of up to 40\% up to the sequence length range of 21-25. Beyond this, as the quantity of training data declines, so does the model's performance. These findings suggest that the DenseBAM-GI model excels in recognizing both complex long-range expressions and short-sequence expressions. The model could perform even better with more data for sequence lengths exceeding 25.

\subsection{Comparative analysis of Computational Complexity of DenseBAM-GI}

Compared to other current, state-of-the-art models in the field, our suggested model, as shown in Table 8, has a lower computational complexity (CC). 
The Computational Complexity (CC) of the proposed DenseBAM-GI can be represented as $(n(d^2 +c') + n(C^2 + (h*w)) + knd^2)$, by using framework established by Vaswani et al. (\cite{Attention_is_all_you_need}).Where, the variables $d$ denote the dimension of the representation, $k$ denotes the size of the convolution, and $n$ is the length of the sequence. In a dense block-generated feature map, the terms $C$ stand for the channels, while $h$ and $w$ stand for height and width, respectively. For adding the additional gate in GRU, the constant $c^{'}$ is used. 

DenseBAM-GI uses a $d$ of 256 and a $n$ range of 0 to 48. The proposed DenseBAM-GI model gets linear and constant calculation time from the auxiliary state $v_t$. With regard to PAL-V2, the CC is roughly $O(knd*nd*nd + knd*2)$. The WAP model has a CC of $O(5(nd2 + knd2))$ since it consists of five different models. The complexity of the Transformer model is around $O(knd2 + 2n2d)$. While the variable $d$ is not overtly defined, it can be noted that in the work of Vaswani et al. (\cite{Attention_is_all_you_need}), this parameter is established as 1000. Our proposed approach only adds a linear polynomial computational time over the base model. In terms of memory allocation, DenseBAM-GI and the base model share similar characteristics. However, DenseBAM-GI exhibits a faster convergence rate, reaching optimal performance within 150 epochs, while only utilizing 6 GB of memory on a 2080 Ti graphics card. In stark contrast, models like Transformer and PAL-V2 necessitate four Nvidia GPUs (either 2080 Ti or 1080 Ti), each contributing 11 GB of memory. Evaluating these data, it is apparent that our proposed DenseBAM-GI model achieves a performance metric equivalent to the state-of-the-art models, yet it does so with a significantly reduced memory footprint and computational requirements, and within a shorter time span.

\section{Conclusion}

This study proposes a novel encoder-decoder architecture (DenseBAM-GI) that address the HMER challenge. The DenseBAM architecture, which includes both the channel and spatial attention mechanisms, is used in our proposed encoder. In addition, we propose the GI-GRU, a novel GRU unit designed to capture, improve, and manage lengthy and complicated expressions as decoder unit. The proposed DenseBAM-GI model, according to experimental results, performs on par with current state-of-the-art models, setting new benchmarks for top 1, 2, and 3 error accuracy while using less processing and memory power. Furthermore, it achieves a stat-of-the art results on CROHME 2019 dataset in terms of exprate. Future research can use this model for extending to other fields, like document recognition and handwritten text recognition.

\section*{Acknowledgement}

We wish to extend our gratitude to the Indian Institute of Information Technology, Allahabad, for supplying the essential research infrastructure that facilitated the execution of this study.


\clearpage

\printbibliography

@inproceedings{10.1145/2402536.2402585_anderson,
author = {Anderson, Robert H.},
title = {Syntax-Directed Recognition of Hand-Printed Two-Dimensional Mathematics},
year = {1967},
isbn = {9781450373098},
publisher = {Association for Computing Machinery},
address = {New York, NY, USA},
url = {https://doi.org/10.1145/2402536.2402585},
doi = {10.1145/2402536.2402585},
booktitle = {Symposium on Interactive Systems for Experimental Applied Mathematics: Proceedings of the Association for Computing Machinery Inc. Symposium},
pages = {436–459},
numpages = {24},
location = {Washington, D.C.}
}

@ARTICLE{CROHME2019,
author = {Mahdavi, Mahshad and Zanibbi, Richard and Mouchère, Harold and Viard-Gaudin, Christian and Garain, Utpal},
year = {2019},
month = {09},
pages = {1533-1538},
title = {ICDAR 2019 CROHME + TFD: Competition on Recognition of Handwritten Mathematical Expressions and Typeset Formula Detection},
doi = {10.1109/ICDAR.2019.00247}
}

@ARTICLE{LSTM,
  author={Hochreiter, Sepp and Schmidhuber, Jürgen},
  journal={Neural Computation}, 
  title={Long Short-Term Memory}, 
  year={1997},
  volume={9},
  number={8},
  pages={1735-1780},
  doi={10.1162/neco.1997.9.8.1735}}

@article{RNN_gradient,
  author={Bengio, Y. and Simard, P. and Frasconi, P.},
  journal={IEEE Transactions on Neural Networks}, 
  title={Learning long-term dependencies with gradient descent is difficult}, 
  year={1994},
  volume={5},
  number={2},
  pages={157-166},
  doi={10.1109/72.279181}}

@ARTICLE{Zanibbi,
	author = {Zanibbi, Richard and Blostein, Dorothea and Cordy, James},
	year = {2002},
	month = {12},
	pages = {1455- 1467},
	title = {Recognizing mathematical expressions using tree transformation. IEEE Trans Pattern Anal Mach Intell},
	volume = {24},
	journal = {Pattern Analysis and Machine Intelligence, IEEE Transactions on},
	doi = {10.1109/TPAMI.2002.1046157}
}

@ARTICLE{Garain,
	author = {Garain, Utpal and Chaudhuri, Bidyut},
	year = {2005},
	month = {01},
	pages = {2366-76},
	title = {Recognition of Online Handwritten Mathematical Expressions},
	volume = {34},
	journal = {IEEE transactions on systems, man, and cybernetics. Part B, Cybernetics : a publication of the IEEE Systems, Man, and Cybernetics Society},
	doi = {10.1109/TSMCB.2004.836817}
}

@inproceedings{yamamoto:inria-00104743,
  TITLE = {{On-Line Recognition of Handwritten Mathematical Expressions Based on Stroke-Based Stochastic Context-Free Grammar}},
  AUTHOR = {Yamamoto, Ryo and Sako, Shinji and Nishimoto, Takuya and Sagayama, Shigeki},
  URL = {https://hal.inria.fr/inria-00104743},
  NOTE = {http://www.suvisoft.com},
  BOOKTITLE = {{Tenth International Workshop on Frontiers in Handwriting Recognition}},
  ADDRESS = {La Baule (France)},
  ORGANIZATION = {{Universit{\'e} de Rennes 1}},
  EDITOR = {Guy Lorette},
  PUBLISHER = {{Suvisoft}},
  YEAR = {2006},
  MONTH = Oct,
  HAL_ID = {inria-00104743},
  HAL_VERSION = {v1},
}

@article{DBLP:journals/corr/RNN_vanishing_gradient,
  author       = {Razvan Pascanu and
                  Tom{\'{a}}s Mikolov and
                  Yoshua Bengio},
  title        = {Understanding the exploding gradient problem},
  journal      = {CoRR},
  volume       = {abs/1211.5063},
  year         = {2012},
  url          = {http://arxiv.org/abs/1211.5063},
  eprinttype    = {arXiv},
  eprint       = {1211.5063},
  bibsource    = {dblp computer science bibliography, https://dblp.org}
}

@ARTICLE{Alvaro1,
	author = {{\'A}lvaro, Francisco and Sánchez, Joan-Andreu and Benedí, José-Miguel},
	year = {2014},
	month = {01},
	pages = {},
	title = {Recognition of on-line handwritten mathematical expressions using 2D stochastic context-free grammars and hidden Markov models},
	volume = {35},
	journal = {Pattern Recognition Letters},
	doi = {10.1016/j.patrec.2012.09.023}
}

@ARTICLE{cho,
	author = {Cho, Kyunghyun and van Merriënboer, Bart and Gulcehre, Caglar and Bougares, Fethi and Schwenk, Holger and Bengio, Y.},
	year = {2014},
	month = {06},
	pages = {},
	title = {Learning Phrase Representations using RNN Encoder-Decoder for Statistical Machine Translation},
	doi = {10.3115/v1/D14-1179}
}

@ARTICLE{crohme2014,
	author = {Mouchère, Harold and Viard-Gaudin, Christian and Zanibbi, Richard and Garain, Utpal},
	year = {2014},
	month = {09},
	pages = {},
	title = {ICFHR 2014 Competition on Recognition of On-line Handwritten Mathematical Expressions (CROHME 2014)},
	volume = {2014},
	journal = {Proceedings of International Conference on Frontiers in Handwriting Recognition, ICFHR},
	doi = {10.1109/ICFHR.2014.42}
}

@ARTICLE{GRU,
	author    = {Junyoung Chung and
	{\c{C}}aglar G{\"{u}}l{\c{c}}ehre and
	KyungHyun Cho and
	Yoshua Bengio},
	title     = {Empirical Evaluation of Gated Recurrent Neural Networks on Sequence
	Modeling},
	journal   = {CoRR},
	volume    = {abs/1412.3555},
	year      = {2014},
	url       = {http://arxiv.org/abs/1412.3555},
	archivePrefix = {arXiv},
	eprint    = {1412.3555},
	bibsource = {dblp computer science bibliography, https://dblp.org}
}

@article{DBLP:journals/corr/BahdanauCB14,
  author    = {Dzmitry Bahdanau and
               Kyunghyun Cho and
               Yoshua Bengio},
  editor    = {Yoshua Bengio and
               Yann LeCun},
  title     = {Neural Machine Translation by Jointly Learning to Align and Translate},
  booktitle = {3rd International Conference on Learning Representations, {ICLR} 2015,
               San Diego, CA, USA, May 7-9, 2015, Conference Track Proceedings},
  year      = {2015},
  url       = {http://arxiv.org/abs/1409.0473},
  bibsource = {dblp computer science bibliography, https://dblp.org}
}

@article{DBLP:journals/corr/cov-additive,
  author    = {Jan Chorowski and
               Dzmitry Bahdanau and
               Dmitriy Serdyuk and
               KyungHyun Cho and
               Yoshua Bengio},
  title     = {Attention-Based Models for Speech Recognition},
  journal   = {CoRR},
  volume    = {abs/1506.07503},
  year      = {2015},
  url       = {http://arxiv.org/abs/1506.07503},
  eprinttype = {arXiv},
  eprint    = {1506.07503},
  bibsource = {dblp computer science bibliography, https://dblp.org}
}

@ARTICLE{Alvaro2,
	author = {{\'A}lvaro, Francisco and Sánchez, Joan-Andreu and Benedí, José-Miguel},
	year = {2015},
	month = {09},
	pages = {},
	title = {An Integrated Grammar-based Approach for Mathematical Expression Recognition},
	volume = {51},
	journal = {Pattern Recognition},
	doi = {10.1016/j.patcog.2015.09.013}
}

@article{Residual_Attention,
  author       = {Fei Wang and
                  Mengqing Jiang and
                  Chen Qian and
                  Shuo Yang and
                  Cheng Li and
                  Honggang Zhang and
                  Xiaogang Wang and
                  Xiaoou Tang},
  title        = {Residual Attention Network for Image Classification},
  journal      = {CoRR},
  volume       = {abs/1704.06904},
  year         = {2017},
  url          = {http://arxiv.org/abs/1704.06904},
  eprinttype    = {arXiv},
  eprint       = {1704.06904},
  bibsource    = {dblp computer science bibliography, https://dblp.org}
}

@inproceedings{luong-etal-2015-effective,
    title = "Effective Approaches to Attention-based Neural Machine Translation",
    author = "Luong, Thang  and
      Pham, Hieu  and
      Manning, Christopher D.",
    booktitle = "Proceedings of the 2015 Conference on Empirical Methods in Natural Language Processing",
    month = sep,
    year = "2015",
    address = "Lisbon, Portugal",
    publisher = "Association for Computational Linguistics",
    url = "https://aclanthology.org/D15-1166",
    doi = "10.18653/v1/D15-1166",
    pages = "1412--1421",
}

@article{wideResNet,
  author       = {Sergey Zagoruyko and
                  Nikos Komodakis},
  title        = {Wide Residual Networks},
  journal      = {CoRR},
  volume       = {abs/1605.07146},
  year         = {2016},
  url          = {http://arxiv.org/abs/1605.07146},
  eprinttype    = {arXiv},
  eprint       = {1605.07146},
  bibsource    = {dblp computer science bibliography, https://dblp.org}
}

@ARTICLE{DBLP:journals/corr/DengKR16,
	author    = {Yuntian Deng and
	Anssi Kanervisto and
	Alexander M. Rush},
	title     = {What You Get Is What You See: {A} Visual Markup Decompiler},
	journal   = {CoRR},
	volume    = {abs/1609.04938},
	year      = {2016},
	url       = {http://arxiv.org/abs/1609.04938},
	archivePrefix = {arXiv},
	eprint    = {1609.04938},
	bibsource = {dblp computer science bibliography, https://dblp.org}
}

@ARTICLE{crohme2016,
	author = {Mouchère, Harold and Viard-Gaudin, Christian and Zanibbi, Richard and Garain, Utpal},
	year = {2016},
	month = {10},
	pages = {607-612},
	title = {ICFHR2016 CROHME: Competition on Recognition of Online Handwritten Mathematical Expressions},
	doi = {10.1109/ICFHR.2016.0116}
}

@article{shattered_gradients,
  author       = {David Balduzzi and
                  Marcus Frean and
                  Lennox Leary and
                  J. P. Lewis and
                  Kurt Wan{-}Duo Ma and
                  Brian McWilliams},
  title        = {The Shattered Gradients Problem: If resnets are the answer, then what
                  is the question?},
  journal      = {CoRR},
  volume       = {abs/1702.08591},
  year         = {2017},
  url          = {http://arxiv.org/abs/1702.08591},
  eprinttype    = {arXiv},
  eprint       = {1702.08591},
  bibsource    = {dblp computer science bibliography, https://dblp.org}
}

@ARTICLE{End-to-End-2017,

	author = {Le Duc, Anh and Nakagawa, Masaki},

	year = {2017},

	month = {11},

	pages = {},

	title = {Training an End-to-End System for Handwritten Mathematical Expression Recognition by Generated Patterns},

	doi = {10.1109/ICDAR.2017.175}

}

@ARTICLE{Densenet,
	author = {Huang, Gao and Liu, Zhuang and van der Maaten, Laurens and Weinberger, Kilian},
	year = {2017},
	month = {07},
	pages = {},
	title = {Densely Connected Convolutional Networks},
	doi = {10.1109/CVPR.2017.243}
}

@inproceedings{Attention_is_all_you_need,
 author = {Vaswani, Ashish and Shazeer, Noam and Parmar, Niki and Uszkoreit, Jakob and Jones, Llion and Gomez, Aidan N and Kaiser, \L ukasz and Polosukhin, Illia},
 booktitle = {Advances in Neural Information Processing Systems},
 editor = {I. Guyon and U. V. Luxburg and S. Bengio and H. Wallach and R. Fergus and S. Vishwanathan and R. Garnett},
 pages = {},
 publisher = {Curran Associates, Inc.},
 title = {Attention is All you Need},
 url = {https://proceedings.neurips.cc/paper/2017/file/3f5ee243547dee91fbd053c1c4a845aa-Paper.pdf},
 volume = {30},
 year = {2017}
}

@ARTICLE{zhang2017watch,
  title={Watch, attend and parse: An end-to-end neural network based approach to handwritten mathematical expression recognition},
  author={Zhang, Jianshu and Du, Jun and Zhang, Shiliang and Liu, Dan and Hu, Yulong and Hu, Jinshui and Wei, Si and Dai, Lirong},
  journal={Pattern Recognition},
  volume={71},
  pages={196--206},
  year={2017},
  publisher={Elsevier}
}

@inproceedings{zhang2018multi,
	title={Multi-scale attention with dense encoder for handwritten mathematical expression recognition},
	author={Zhang, Jianshu and Du, Jun and Dai, Lirong},
	booktitle={International Conference on Pattern Recognition},
	pages={2245--2250},
	year={2018}
}

@article{DBLP:journals/corr/BAM,
  author       = {Jongchan Park and
                  Sanghyun Woo and
                  Joon{-}Young Lee and
                  In So Kweon},
  title        = {{BAM:} Bottleneck Attention Module},
  journal      = {CoRR},
  volume       = {abs/1807.06514},
  year         = {2018},
  url          = {http://arxiv.org/abs/1807.06514},
  eprinttype    = {arXiv},
  eprint       = {1807.06514},
  bibsource    = {dblp computer science bibliography, https://dblp.org}
}

@ARTICLE{inbook,
	author = {Wu, Jin-Wen and Yin, Fei and Zhang, Yan-Ming and Zhang, Xu-Yao and Liu, Cheng-Lin},
	year = {2019},
	month = {01},
	pages = {18-34},
	title = {Image-to-Markup Generation via Paired Adversarial Learning: Recognizing Outstanding Ph.D. Research},
	isbn = {978-981-13-6048-0},
	doi = {10.1007/978-3-030-10925-7_2}
}

@ARTICLE{multimodal,
	author = {Wang, Jiaming and Du, Jun and Zhang, Jianshu and Wang, Zi-Rui},
	year = {2019},
	month = {09},
	pages = {1181-1186},
	title = {Multi-modal Attention Network for Handwritten Mathematical Expression Recognition},
	doi = {10.1109/ICDAR.2019.00191}
}

@InProceedings{DenseWAP-TD,
  title = 	 {A Tree-Structured Decoder for Image-to-Markup Generation},
  author =       {Zhang, Jianshu and Du, Jun and Yang, Yongxin and Song, Yi-Zhe and Wei, Si and Dai, Lirong},
  booktitle = 	 {Proceedings of the 37th International Conference on Machine Learning},
  pages = 	 {11076--11085},
  year = 	 {2020},
  editor = 	 {III, Hal Daumé and Singh, Aarti},
  volume = 	 {119},
  series = 	 {Proceedings of Machine Learning Research},
  month = 	 {13--18 Jul},
  publisher =    {PMLR},
  url = 	 {https://proceedings.mlr.press/v119/zhang20g.html}
}

@ARTICLE{dual-loss,

	author = {Le Duc, Anh},

	year = {2020},

	month = {06},

	pages = {2413-2418},

	title = {Recognizing handwritten mathematical expressions via paired dual loss attention network and printed mathematical expressions},

	doi = {10.1109/CVPRW50498.2020.00291}

}

@ARTICLE{PAL-V2,
	author = {Wu, Jin-Wen and Yin, Fei and Zhang, Yan-Ming and Zhang, Xu-Yao and Liu, Cheng-Lin},
	year = {2020},
	month = {01},
	pages = {},
	title = {Handwritten Mathematical Expression Recognition via Paired Adversarial Learning},
	journal = {International Journal of Computer Vision},
	doi = {10.1007/s11263-020-01291-5}
}

@ARTICLE{DBLP:journals/corr/abs-2006-06919_momentum_rnn,
  author    = {Tan M. Nguyen and
               Richard G. Baraniuk and
               Andrea L. Bertozzi and
               Stanley J. Osher and
               Bao Wang},
  title     = {MomentumRNN: Integrating Momentum into Recurrent Neural Networks},
  journal   = {CoRR},
  volume    = {abs/2006.06919},
  year      = {2020},
  url       = {https://arxiv.org/abs/2006.06919},
  eprinttype = {arXiv},
  eprint    = {2006.06919},
  bibsource = {dblp computer science bibliography, https://dblp.org}
}

@inproceedings{DenseWAP_CTC,
author = {Nguyen, Cuong Tuan and Nguyen, Hung Tuan and Morizumi, Kei and Nakagawa, Masaki},
title = {Temporal Classification Constraint for Improving Handwritten Mathematical Expression Recognition},
year = {2021},
isbn = {978-3-030-86158-2},
publisher = {Springer-Verlag},
address = {Berlin, Heidelberg},
url = {https://doi.org/10.1007/978-3-030-86159-9_8},
doi = {10.1007/978-3-030-86159-9_8},
booktitle = {Document Analysis and Recognition – ICDAR 2021 Workshops: Lausanne, Switzerland, September 5–10, 2021, Proceedings, Part II},
pages = {113–125},
numpages = {13},
location = {Lausanne, Switzerland}
}

@ARTICLE{DBLP:journals/corr/abs-2105-02412-transformer,
  author    = {Wenqi Zhao and
               Liangcai Gao and
               Zuoyu Yan and
               Shuai Peng and
               Lin Du and
               Ziyin Zhang},
  title     = {Handwritten Mathematical Expression Recognition with Bidirectionally
               Trained Transformer},
  journal   = {CoRR},
  volume    = {abs/2105.02412},
  year      = {2021},
  url       = {https://arxiv.org/abs/2105.02412},
  eprinttype = {arXiv},
  eprint    = {2105.02412},
  bibsource = {dblp computer science bibliography, https://dblp.org}
}

@Article{R-GRU,
author={Pal, Aniket
and Singh, Krishna Pratap},
title={R-GRU: Regularized gated recurrent unit for handwritten mathematical expression recognition},
journal={Multimedia Tools and Applications},
year={2022},
month={Apr},
day={09},
issn={1573-7721},
doi={10.1007/s11042-022-12889-x},
url={https://doi.org/10.1007/s11042-022-12889-x}
}

@article{resnet,
      title={Deep Residual Learning for Image Recognition}, 
      author={Kaiming He and Xiangyu Zhang and Shaoqing Ren and Jian Sun},
      year={2015},
      eprint={1512.03385},
      archivePrefix={arXiv},
      primaryClass={cs.CV}
}

@article{vggnet,
      title={Very Deep Convolutional Networks for Large-Scale Image Recognition}, 
      author={Karen Simonyan and Andrew Zisserman},
      year={2015},
      eprint={1409.1556},
      archivePrefix={arXiv},
      primaryClass={cs.CV}
}

@article{AdamR-GRU,
title = {AdamR-GRUs: Adaptive momentum-based Regularized GRU for HMER problems},
journal = {Applied Soft Computing},
pages = {110457},
year = {2023},
issn = {1568-4946},
doi = {https://doi.org/10.1016/j.asoc.2023.110457},
url = {https://www.sciencedirect.com/science/article/pii/S1568494623004751},
author = {Aniket Pal and Krishna Pratap Singh}
}





\end{document}